\documentclass[journal]{IEEEtran}
\usepackage{amsmath,amssymb}
\usepackage{algorithm}
\usepackage{algpseudocode}
\usepackage{graphicx}
\usepackage{booktabs}
\usepackage{multirow}
\usepackage{color}
\usepackage{url}
\usepackage[hidelinks]{hyperref}

\graphicspath{{figures/}}

\newcommand{\best}[1]{\textbf{#1}}
\newcommand{\R}{\mathbb{R}}

\begin{document}

\title{S2WMamba: Modality-Specific Wavelet Decomposition with Learnable Detail
Bands for Pansharpening}

\author{Haoyu~Zhang$^{\ast}$, Junhan~Luo$^{\ast}$, Yugang~Cao$^{\ast}$, and
Liang-Jian~Deng%
\thanks{Manuscript received November 2026.}%
\thanks{The authors are with the School of Mathematical Sciences, University of
Electronic Science and Technology of China, Chengdu 611731, China~(e\mbox{-}mail:
2024090908014@std.uestc.edu.cn; 2024310207017@std.uestc.edu.cn;
2024080301024@std.uestc.edu.cn; liangjian.deng@uestc.edu.cn).}%
\thanks{$^{\ast}$These authors contributed equally to this work. Corresponding
author: Liang-Jian Deng.}}

\markboth{IEEE Transactions on Geoscience and Remote Sensing}{Zhang \emph{et al.}: S2WMamba}

\maketitle

\begin{abstract}
Pansharpening reconstructs a high-resolution multispectral image from a
high-resolution panchromatic (PAN) image and a low-resolution multispectral (MS)
image. The MS bands carry the colour that must be preserved, but they enter most
networks as an undifferentiated feature dimension, mixed with the spatial
correction before any spectral decomposition is formed. We propose S2WMamba
(spatial, spectral, wavelet, Mamba), which treats the spatial and the spectral axes
with different transforms: a
two-dimensional wavelet decomposition applied to a learned embedding of the PAN
image isolates spatial edges and textures, while a one-dimensional wavelet
decomposition applied along the spectral axis of the fused cue separates the
spectral base shared by all bands from the band-specific
variation. In both transforms the approximation (low-pass) filter is
\emph{fixed} and carries no learned parameters, and only the detail filters are
learned.
The two branches are coupled in one direction: the spatial branch first
reconstructs a correction to the upsampled MS image, which then conditions the
input of the spectral branch. A Mamba-based aggregator merges the two branch outputs. On the
PanCollection benchmark, S2WMamba attains the best reduced-resolution PSNR on
each of WorldView-3 (39.502~dB), GF2 (50.083~dB) and QuickBird (38.802~dB), and
in full resolution its HQNR is within 0.011 of the best on every sensor. Ablations show that the benefit of the cascade lies entirely in its
\emph{direction}: reversing it yields the same accuracy as removing the coupling
altogether. Code: \mbox{\url{github.com/KagUYa66/S2WMamba}}; benchmark:
\mbox{\url{github.com/liangjiandeng/PanCollection}}.
\end{abstract}


\section{Introduction}
\IEEEPARstart{P}{ansharpening} fuses a panchromatic (PAN) image, which has high
spatial resolution but a single band, with a multispectral (MS) image, which has
rich spectral content but a coarse footprint, to produce a high-resolution
multispectral (HRMS) product. It is a standard preprocessing step for
classification, change detection and mapping, and it is conventionally evaluated
either against a spatially degraded reference (reduced resolution) or with
no-reference indices (full resolution).

\begin{figure}[!t]
\centering
\includegraphics[width=\columnwidth]{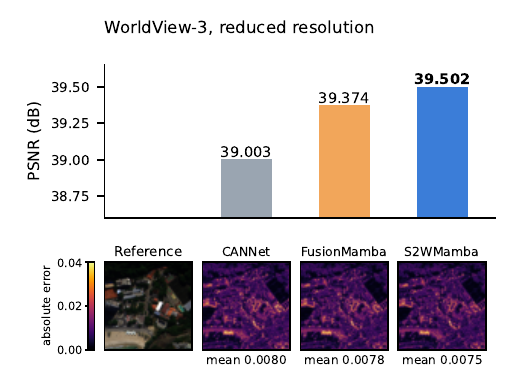}
\caption{Reduced-resolution results on WorldView-3. Top: PSNR over the 20 test
images for two strong baselines and S2WMamba. Bottom: one test image with the
absolute error of each method against the reference, on a common colour scale; the
mean error is given under each method panel.}
\label{fig:teaser}
\end{figure}

Two families of methods have dominated. Classical component-substitution and
multiresolution schemes rely on a hand-designed injection rule, usually derived
from the modulation transfer function of the sensor, and their spectral fidelity
degrades quickly when the injection gain is misestimated
\cite{vivone2015critical,vivone2018multiresolution,palsson2013variational}. Deep
networks replaced that rule with a learned residual: from early convolutional
models to the recent U-shaped and attention-based designs
\cite{masi2016pansharpening,yang2017pannet,he2019dicnn,deng2021fusionnet,duan2024cannet,peng2023u2net},
the dominant pattern is to upsample the MS image and add a high-frequency residual
predicted from the PAN image. More recent work brings state space models into the
fusion stage \cite{he2024panmamba,peng2024fusionmamba}, gaining long-range
modelling at linear cost.

\emph{What is still missing.} In these designs the MS bands enter as an
undifferentiated feature dimension: a convolution or an attention map over all
$C$ bands mixes the spectral base with band-specific variation before the
correction is formed. Yet the two axes call
for different decompositions. The PAN image is a spatial signal whose informative
content is concentrated in edges and textures, which a two-dimensional transform
isolates. The MS image carries a spectrum per pixel; its bands are strongly
correlated, and the variation that must be preserved is the deviation of each band
from the spectral base shared by the others. Treating the channel axis of the MS
tensor as if it were another spatial axis discards that structure, and treating it
as an unordered set discards the fact that adjacent bands are physically
neighbouring wavelengths.

A second issue is what a learned decomposition is allowed to do. Replacing the
wavelet filters by free convolutions is the obvious way to gain adaptivity, but it
is not what the data support: the global residual path already carries the low
frequencies of the MS image, so what a learned low-pass changes is not the
low-frequency content of the output but the meaning of the decomposition, and
freeing it costs 0.024~dB rather than gaining. Our experiments indicate that the
useful ingredient is narrower than ``learn the whole basis''.

\emph{This paper.} S2WMamba applies a \emph{modality-specific} pair of
decompositions and constrains them differently from prior work. First, the PAN
image is decomposed by a two-dimensional wavelet transform, and the fused cue is
decomposed by a one-dimensional wavelet transform applied along the channel axis. Second, in both transforms the approximation band is produced by a
\emph{fixed} low-pass filter --- the $2\times2$ average for the PAN and a fixed
low-pass for the spectral axis --- while the detail bands are learned. The
fixed branch carries no parameters; the learned branch is free to reweight
detail per level, which is where our ablations locate most of the gain over
FusionMamba (0.124 of 0.128~dB). Third, the two branches are
cascaded rather than run in parallel: the spatial branch produces a correction,
and its output is added to the upsampled MS image to form the cue from which the
spectral branch's channel sub-bands are computed. We show by ablation that this
direction is not a detail: reversing it gives the same accuracy as removing the
coupling altogether, i.e. the benefit comes from letting the spatial branch
condition the spectral branch, not from cross-branch communication in general.
Fig.~\ref{fig:teaser} summarizes the reduced-resolution comparison on WorldView-3.

Our contributions are summarised as follows.
\begin{itemize}
\item A modality-specific wavelet decomposition for pansharpening: a 2-D transform
of a learned embedding of the PAN image for spatial structure, and a 1-D transform
along the spectral axis of the fused cue for the spectral base, so that the two axes
are handled by separate operators instead of one shared convolution.
\item A constrained parameterisation in which only the detail bands are learnable.
We show that a fully learned basis is \emph{worse} than the fixed Haar basis in
PSNR and ERGAS despite a larger parameter count, and that additionally fixing
every detail filter to unit norm also loses accuracy; the gain therefore comes
from per-level gain freedom rather than from a fixed filter norm.
\item A directed cascade whose value is established by ablation against both its
reverse and its parallel counterpart at identical parameter count.
\item Best reduced-resolution PSNR under an identical protocol on all three sensors:
+0.128~dB over FusionMamba on WorldView-3, +0.174~dB over ADWM on GF2 and
+0.314~dB over CANNet on QuickBird.
\end{itemize}

\begin{figure*}[!t]
\centering
\includegraphics[width=0.90\textwidth]{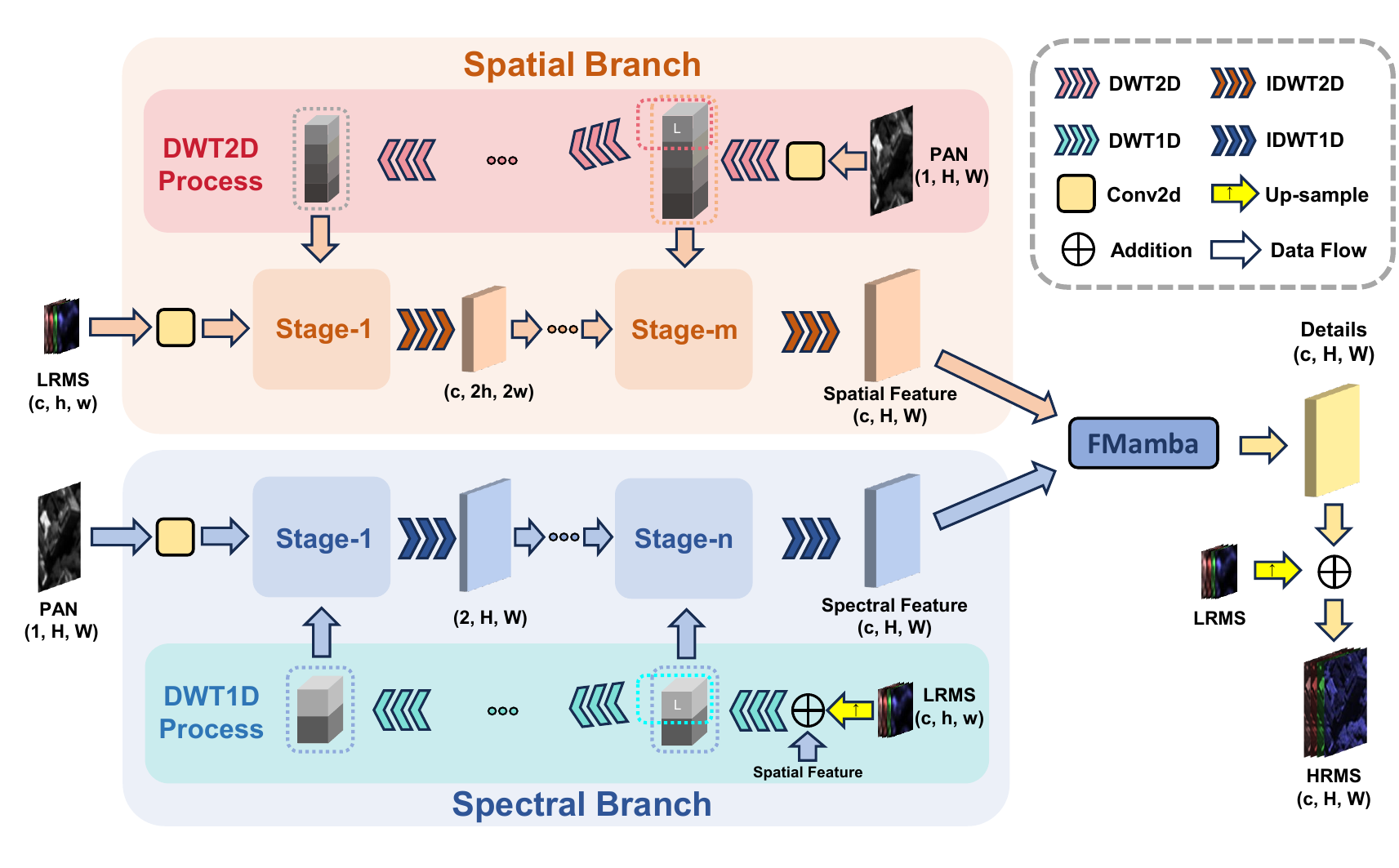}
\caption{Overview of S2WMamba. The PAN image is embedded by a $3\times3$ convolution
and decomposed by a 2-D wavelet transform whose approximation filter is a fixed
$2\times2$ average and whose detail filters are learned; the fused cue --- the spatial
branch's output added to the upsampled MS image --- is decomposed by a 1-D transform
along the channel axis with the same fixed/learned split. The 2-D sub-bands condition
the spatial branch and the 1-D sub-bands the spectral branch; the two outputs are
merged by the FM aggregator and added to the upsampled MS image.}
\label{fig:arch}
\end{figure*}

\section{Related Work}
\subsection{Pansharpening}
Component-substitution methods project the MS image into a space where the PAN
image can replace one component, and differ in how the projection and the
histogram matching are defined. Multiresolution-analysis methods instead inject
the high-frequency content of the PAN image at a scale determined by the sensor's
modulation transfer function \cite{vivone2018multiresolution};
MTF-based variants remain strong baselines that are hard to beat on spectral
fidelity, and the full-resolution assessment of such schemes has been studied in
detail \cite{arienzo2022full}. Variational formulations cast fusion as the
minimisation of a data term plus spatial and spectral priors
\cite{palsson2013variational}.

Deep pansharpening began with PNN \cite{masi2016pansharpening} and PanNet
\cite{yang2017pannet}, which learn the residual between the upsampled MS image and
the reference, and continued with lighter and deeper convolutional variants such as
DiCNN \cite{he2019dicnn}. FusionNet \cite{deng2021fusionnet} introduced a
two-stream residual design with full-resolution supervision, and DCFNet
\cite{wu2021dcfnet} fused features across two branches at every scale. LAGConv
\cite{jin2022lagconv} made the convolution kernel itself content-adaptive.
CANNet \cite{duan2024cannet} made the convolution content-adaptive and non-local. Transformer
backbones have since been explored in several forms
\cite{meng2022vit,quan2023dtempan,wang2023duald,zhao2023asym,su2022trn,zhang2022amf},
of which PanFormer \cite{zhou2022panformer} is a representative, and U-shaped
convolutional designs have been revisited in U2Net \cite{peng2023u2net}.
Back-projection has been used to couple the two
streams \cite{zhang2023bp}, quaternion algebra to tie the spectral bands together
\cite{li2025quaternion}, and zero-shot variational inference has been revisited for
the multispectral case \cite{rui2024zero}. Two further lines address the training
signal rather than the architecture: semi-supervised learning across resolutions
\cite{yang2023crossres} and spatial data augmentation for generalisation
\cite{chen2023aug}. Pan-Mamba \cite{he2024panmamba}
replaced the fusion stage with a state space model. FusionMamba \cite{peng2024fusionmamba} couples two Mamba streams
through a cross-modulation module and is the strongest published baseline on
WorldView-3 in our comparison. A recent line replaces the fusion operator
by a diffusion process \cite{xing2024crossdiff,zhong2024ssdiff}, or by dual-level
weighting (ADWM) \cite{huang2025adwm}. Our work differs from all of them in that the
\emph{decomposition itself} is matched to the modality and is constrained, rather
than in the choice of the fusion operator alone.

\subsection{Wavelet-based Fusion}
Wavelet fusion injects the PAN detail coefficients into the MS bands and
reconstructs with the inverse transform \cite{mallat1989theory}; the additive
wavelet luminance proportional rule (AWLP) is the reference implementation of this
idea and a standard baseline in the field \cite{vivone2015critical}. Recent
segmentation networks carry the decomposition inside the network and fuse jointly
in the spatial and frequency domains \cite{yang2024sffnet}, an idea also pursued in
image restoration \cite{cui2024adair}. The classical weakness of
wavelet fusion is that a fixed basis cannot adapt to the sensor or the scene. A
fully learned transform frees
every filter, including the one that produces the approximation band. We keep the
approximation path fixed for both modalities and learn only the detail filters,
which preserves the interpretation of the decomposition while retaining the
adaptivity that the fixed basis lacks.

\subsection{State Space Models}
Selective state space models scan a sequence with input-dependent transition
parameters, giving long-range dependencies at linear cost \cite{gu2023mamba}. In
vision they have been arranged as multi-directional scans over flattened feature
maps \cite{liu2024vmamba,baty2024mamba2d}, which is the form we use inside both
branches and inside the aggregator.

For pansharpening the appeal of this formulation is specific. The decision that
matters at a given pixel --- how much PAN detail to inject and in which spectral
band --- depends on a neighbourhood whose extent is set by the scene rather than by the
kernel size: a building edge and a coastline need different support.
Convolutional fusion blocks fix that extent in advance, and attention-based blocks
pay a quadratic cost to let it vary. A selective scan makes the transition
parameters a function of the input, so the effective support adapts, at a cost that
remains linear in the number of pixels. Both branches use such a scan, on
different sequences: the spatial branch over the decomposed PAN structure, the
spectral branch over the channel-decomposed MS estimate --- one operator serving
two modalities whose informative content is not the same kind.

\begin{figure*}[!t]
\centering
\includegraphics[width=0.90\textwidth]{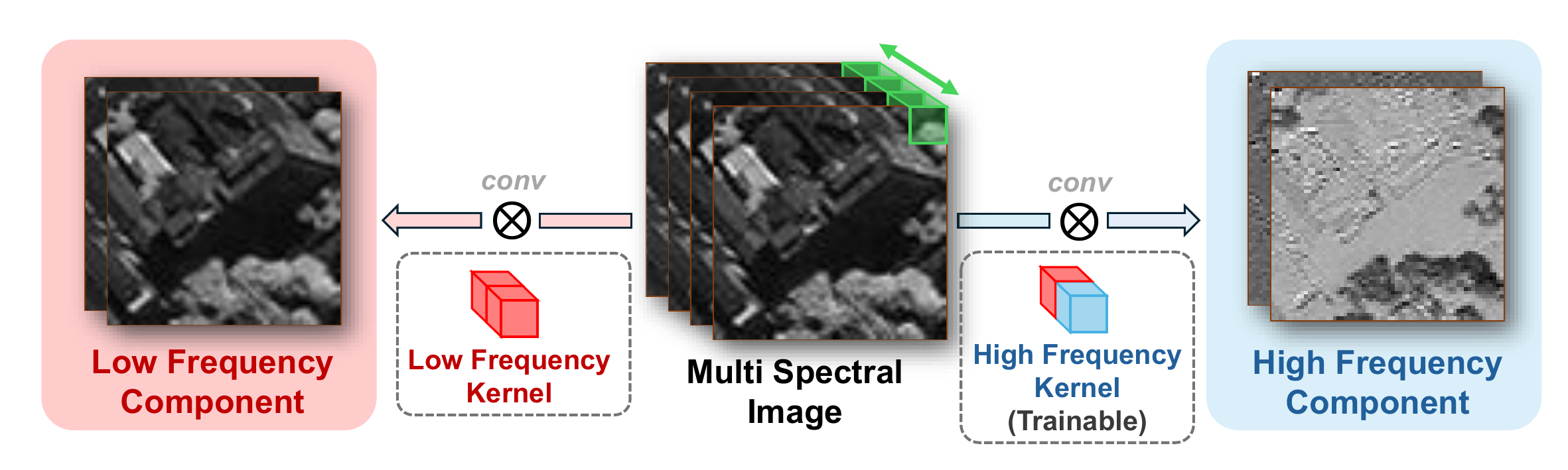}
\caption{The constrained decomposition, drawn for the channel axis. The
approximation component is produced by a fixed low-frequency kernel,
the detail component by a learnable high-frequency kernel; only the latter is
updated during training.}
\label{fig:dwt}
\end{figure*}

\section{Method}
\subsection{Overview}
Let $\mathbf{M}\in\R^{C\times H/4\times W/4}$ be the MS image and
$\mathbf{P}\in\R^{1\times H\times W}$ the PAN image. The overall structure is shown
in Fig.~\ref{fig:arch}. The network keeps a global residual path, so the output is
\begin{equation}
\hat{\mathbf{Y}} = \uparrow_4(\mathbf{M}) + \mathcal{G}(\mathbf{F}_{sp},\mathbf{F}_{se}),
\label{eq:global}
\end{equation}
where $\uparrow_4$ is bilinear upsampling, $\mathbf{F}_{sp}$ and $\mathbf{F}_{se}$ are the
spatial and spectral branch outputs and $\mathcal{G}$ is the aggregator. The two
branches share no parameters, but they are not independent: the spectral branch is
conditioned on the spatial branch's output, as described in Section~\ref{sec:cascade}
and summarised in Algorithm~\ref{alg:forward}.

\subsection{Modality-Specific Decomposition}
\emph{Spatial axis.} The PAN image is embedded into $C_f=32$ feature maps by a
$3\times3$ convolution $f_{pan}$ and decomposed by a two-level 2-D wavelet
transform,
\begin{equation}
\begin{aligned}
\{\mathbf{A}^{(1)},\mathbf{D}^{(1)}\} &= \mathcal{W}_{2D}(f_{pan}(\mathbf{P})),\\
\{\mathbf{A}^{(2)},\mathbf{D}^{(2)}\} &= \mathcal{W}_{2D}(\mathbf{A}^{(1)}),
\end{aligned}
\label{eq:dwt2d}
\end{equation}
where $\mathbf{A}^{(\ell)}$ is the approximation band of level $\ell$ and
$\mathbf{D}^{(\ell)}=\{\mathbf{H}_1^{(\ell)},\mathbf{H}_2^{(\ell)},\mathbf{H}_3^{(\ell)}\}$
are the three detail bands of that level; each band is a $C_f$-channel map, so the
transform acts on the $C_f$-dimensional embedding and not on the single PAN channel.
Each level thus contributes four bands, and the spatial branch receives all eight
to inject spatial structure into the MS features.

\emph{Spectral axis.} For a tensor $\mathbf{Z}\in\R^{C\times H\times W}$ the
channel-wise transform pairs adjacent bands and applies a two-point wavelet along
the channel axis for $N=\log_2 C$ levels, producing a low-pass and a high-pass half
at each level:
\begin{equation}
\begin{aligned}
\mathbf{L}^{(1)},\mathbf{H}^{(1)} &= \mathcal{W}_{1D}(\mathbf{Z}),\\
\mathbf{L}^{(\ell)},\mathbf{H}^{(\ell)} &= \mathcal{W}_{1D}(\mathbf{L}^{(\ell-1)}),
\quad \ell=2,\dots,N.
\end{aligned}
\end{equation}
$\mathbf{L}^{(\ell)}$ aggregates what neighbouring bands have in common --- the
spectral base --- and $\mathbf{H}^{(\ell)}$ holds what distinguishes them.
$N$ is three for the eight-band WorldView-3 and two for the four-band sensors,
where the second application already leaves a single channel.
The spectral branch is fed with both halves at every level, at the resolution of
the PAN image (Fig.~\ref{fig:dwt}).

The choice of pairing \emph{adjacent} bands is a modelling assumption, and we test
it directly: permuting the input band order of the trained model costs 9.5~dB on
WorldView-3 and 6.3~dB on QuickBird, so physical adjacency, not a channel-wise
transform as such, is what the decomposition exploits.

\subsection{Fixed Approximation, Learnable Detail}
\label{sec:lofix}
A learned two-point transform would replace both rows of the $2\times2$ matrix by
free parameters. We constrain it instead. For the spectral axis the low-pass row is
fixed to the Haar average and only the high-pass row is learnable; for the spatial
axis the approximation band of the level input $\mathbf{X}$ is the fixed $2\times2$
Haar average, scaled by two as in the Haar filter bank,
\begin{equation}
\mathbf{A} = 2\,\mathrm{avgpool}_{2}(\mathbf{X}),\qquad
\mathbf{H}_{\theta} = \mathbf{X} * \mathbf{K}_{\theta}\ \ (\text{stride }2),
\end{equation}
where $\mathbf{K}_{\theta}$ is a $2\times2$ convolution with $3C_f$ output channels,
i.e. one learned kernel per detail band and feature channel. The synthesis step
applies the transposed filters, so it is the adjoint of the analysis rather than
its exact inverse: with a learned detail kernel the four bands no longer satisfy a
perfect-reconstruction identity. We keep the wavelet terminology as shorthand for
this critically sampled analysis--synthesis filter bank. The approximation path
therefore holds no learned
parameters, and the detail bands are the only part of the transform that training
can change.

This parameterisation is deliberately narrower than a free basis, and the
ablations in Section~\ref{sec:ablation} show that the narrowness matters: freeing
all filters lowers PSNR by 0.024~dB relative to the fixed Haar basis, whereas
freeing only the detail bands raises it by 0.124~dB. We also tested a variant in
which each detail filter is renormalised to unit norm at every forward pass, so
that its energy is fixed while its direction stays free; it falls \emph{below} the
fixed Haar basis. The gain of the proposed parameterisation is therefore per-level
gain freedom, not a fixed filter norm.

\begin{figure*}[!t]
\centering
\includegraphics[width=0.86\textwidth]{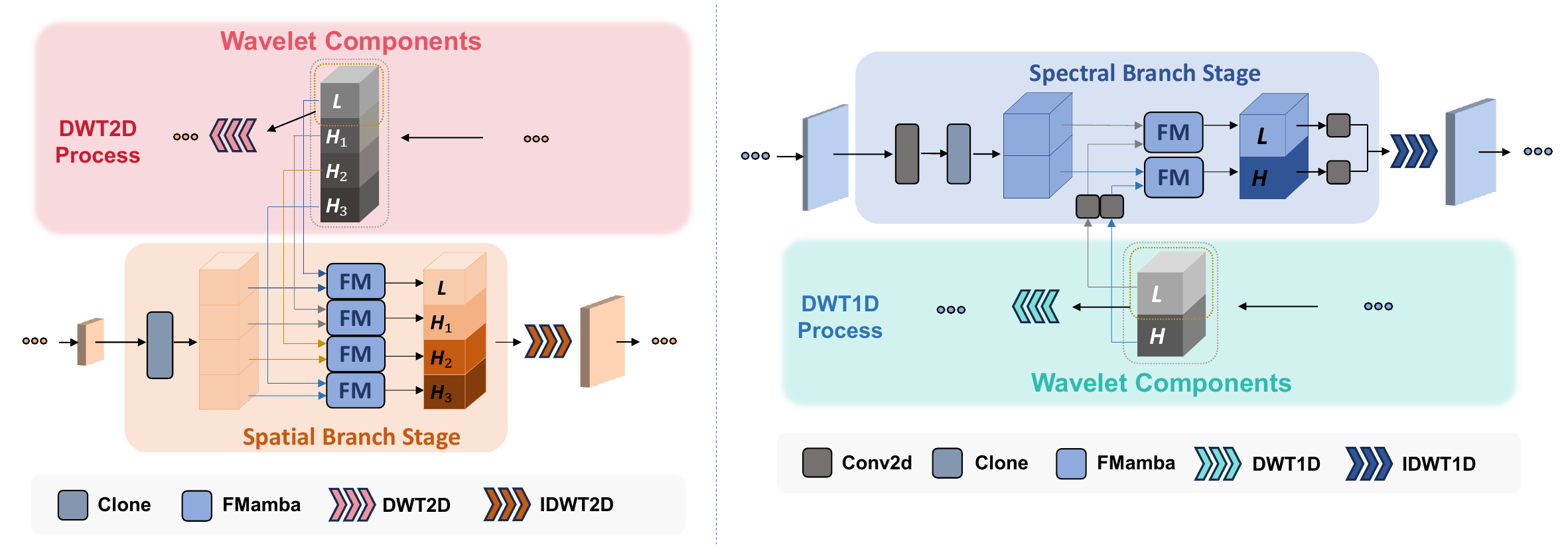}
\caption{One stage of each branch. The spatial branch (left) carries the MS-derived
features and is modulated at each of the four 2-D sub-band positions --- the
approximation band, written $L$ in the figure and $\mathbf{A}^{(\ell)}$ in
\eqref{eq:dwt2d}, and the three detail bands
$\mathbf{H}^{(\ell)}_1,\mathbf{H}^{(\ell)}_2,\mathbf{H}^{(\ell)}_3$ --- then
reconstructs by the 2-D synthesis transform. The spectral branch (right) carries
the PAN-derived features, is modulated by the channel-wise $L$ and $H$ halves, and
reconstructs by the 1-D synthesis transform.}
\label{fig:stages}
\end{figure*}

\begin{algorithm}[!tb]
\caption{S2WMamba forward pass}
\label{alg:forward}
\begin{algorithmic}[1]
\Require PAN $\mathbf{P}\!\in\!\R^{1\times H\times W}$, MS
$\mathbf{M}\!\in\!\R^{C\times H/4\times W/4}$;\ \ $N=\log_2 C$
\Statex $f_{sp},f_{se}$: branch input embeddings; $\mathrm{FM}$: one cross-modal
Mamba block; $\mathcal{G}$: FM aggregator
\State $\mathbf{M}^{\uparrow}\leftarrow\,\uparrow_4\!(\mathbf{M})$
\State $\{\mathbf{A}^{(1)},\mathbf{D}^{(1)}\}\leftarrow\mathcal{W}_{2D}(f_{pan}(\mathbf{P}))$;\ \
$\{\mathbf{A}^{(2)},\mathbf{D}^{(2)}\}\leftarrow\mathcal{W}_{2D}(\mathbf{A}^{(1)})$
\Comment{fixed low-pass, learned detail}
\State $\mathbf{F}_{sp}\leftarrow f_{sp}(\mathbf{M})$
\State $\mathbf{F}_{sp}\leftarrow\mathrm{IDWT}_{2D}\bigl(\{\mathrm{FM}(\mathbf{F}_{sp}\mid s)
: s\in\{\mathbf{A}^{(2)},\mathbf{D}^{(2)}\}\}\bigr)$ \Comment{four bands, four blocks}
\State $\mathbf{F}_{sp}\leftarrow\mathrm{IDWT}_{2D}\bigl(\{\mathrm{FM}(\mathbf{F}_{sp}\mid s)
: s\in\{\mathbf{A}^{(1)},\mathbf{D}^{(1)}\}\}\bigr)$
\State $\mathbf{Z}\leftarrow\mathbf{F}_{sp}+\mathbf{M}^{\uparrow}$
\Comment{cascade cue}
\State $\{\mathbf{L}^{(\ell)},\mathbf{H}^{(\ell)}\}_{\ell=1}^{N}\leftarrow
\mathcal{W}_{1D}(\mathbf{Z})$
\State $\mathbf{F}_{se}\leftarrow f_{se}(\mathbf{P})$
\For{$\ell=N$ \textbf{downto} $1$}
  \State $\mathbf{U}\leftarrow\mathrm{FM}(\mathbf{F}_{se}\mid\mathbf{L}^{(\ell)})$;\ \
  $\mathbf{V}\leftarrow\mathrm{FM}(\mathbf{F}_{se}\mid\mathbf{H}^{(\ell)})$
  \State $\mathbf{F}_{se}\leftarrow\mathrm{IDWT}_{1D}(\mathbf{U},\mathbf{V})$
\EndFor
\State $\mathbf{R}\leftarrow\mathcal{G}(\mathbf{F}_{sp},\mathbf{F}_{se})$
\State \Return $\mathbf{M}^{\uparrow}+\mathbf{R}$
\end{algorithmic}
\end{algorithm}

\subsection{Spatial Branch}
The spatial branch starts at the MS resolution and reconstructs to the PAN
resolution. Its input is $\mathbf{M}$; its side information is the two-level 2-D
decomposition of the PAN image. At each level the four sub-bands are fused with
the MS features by four separate cross-modal blocks, one per sub-band position,
and the results are recombined by the 2-D synthesis transform: the first stage
acts on $\{\mathbf{A}^{(2)},\mathbf{D}^{(2)}\}$, the second on
$\{\mathbf{A}^{(1)},\mathbf{D}^{(1)}\}$, and the final 2-D synthesis transform
produces $\mathbf{F}_{sp}\in\R^{C\times H\times W}$.

The coarser level is injected first, so that the fine level refines an estimate
that already contains the large-scale structure of the PAN image. The block flattens its input into a
sequence and scans it in four directions --- the two raster orders and their
transposes --- with the other branch's features entering the scan as an extra
embedding that modulates the state-space updates; a gated residual is then added.

\subsection{Spectral Branch}
\label{sec:cascade}
The spectral branch operates at the PAN resolution and is the component that makes
the cascade directional. Its side information is the channel-wise decomposition of
a \emph{cue} formed by the spatial branch's output added to the upsampled MS image
(Eq.~\eqref{eq:cue}),
\begin{equation}
\mathbf{Z} = \mathbf{F}_{sp} + \uparrow_4(\mathbf{M}),
\qquad
\{\mathbf{L}^{(\ell)},\mathbf{H}^{(\ell)}\}_{\ell=1}^{N} = \mathcal{W}_{1D}(\mathbf{Z}).
\label{eq:cue}
\end{equation}
$N$ stages then fuse the PAN features with progressively finer spectral
detail (Fig.~\ref{fig:stages}), each stage running two cross-modal Mamba blocks ---
one against the low-pass and one against the high-pass half --- followed by the 1-D
synthesis transform. The output is $\mathbf{F}_{se}$.

Because $\mathbf{Z}$ contains $\mathbf{F}_{sp}$, the spectral branch sees the spatial
branch's correction to the upsampled MS image before injecting detail. Reversing the direction
--- running the spectral branch first on $\uparrow_4(\mathbf{M})$ and adding its
output, downsampled to the MS grid, to $\mathbf{M}$ before the spatial branch ---
is possible with the same modules and the same parameter count, and
Section~\ref{sec:ablation} reports it.

\subsection{Aggregation and Loss}
The two branch outputs are merged by an aggregator built from the same cross-modal
block: a shared $1\times1$ projection lifts each input to the block width, the block
fuses the two projections in both directions with a gated residual, and a $1\times1$
convolution maps back to $C$ channels. The projection is shared so that the
aggregator cannot tell the two inputs apart by their embedding.
The aggregated residual is added to the upsampled MS image as in
Eq.~\eqref{eq:global}.
The network is trained with the $\ell_1$ loss of \eqref{eq:loss} between
$\hat{\mathbf{Y}}$ and the reference HRMS image $\mathbf{Y}$ \cite{zhao2017loss}, as
in the baselines we compare with,
\begin{equation}
\mathcal{L} = \frac{1}{CHW}\sum_{c=1}^{C}\sum_{h=1}^{H}\sum_{w=1}^{W}
\left|\hat{\mathbf{Y}}_{c,h,w}-\mathbf{Y}_{c,h,w}\right| .
\label{eq:loss}
\end{equation}

\section{Experiments}
\subsection{Datasets and Protocols}
All experiments use PanCollection \cite{vivone2019component,deng2022machine}, which
provides training and test splits for WorldView-3 (8 bands), GF2 (4 bands) and
QuickBird (4 bands). Reduced-resolution (RR) evaluation follows Wald's protocol
\cite{wald1997fusion,wald2002data}: the PAN and MS images are spatially degraded by a
factor of four, the network is trained on the degraded pairs, and PSNR, SAM
\cite{boardman1993automated,yuhas1992discrimination}, ERGAS and
$Q_{8}$/$Q_{4}$ are computed against the original MS image as reference, with a
30-pixel border excluded. Full-resolution (FR) evaluation uses the original PAN and
MS images and the no-reference indices \cite{alparone2008qnr} $D_\lambda$, $D_s$
and $\mathrm{HQNR}=(1-D_\lambda)(1-D_s)$, computed with a 32-pixel block size.

\subsection{Implementation Details}
The network is trained for 500 epochs with AdamW \cite{loshchilov2017decoupled}, an
initial learning rate of $4\times10^{-4}$ decayed by a factor of 0.7 every 100
epochs, a batch size of 32, and $64\times64$ PAN patches, in PyTorch. The width
is 32.
Each cross-modal block is a four-direction selective scan with state dimension 8 and
expansion factor 1; the aggregator is a third block of the same kind, at a width of
84 for WorldView-3 and 64 for the four-band sensors. On a single RTX~5090 a WorldView-3
training run takes about 7 hours. Reported S2WMamba weights are the checkpoint with
the best test-split PSNR, used unchanged for both protocols. The choice moves the
headline by at most 0.044~dB: in steps of ten epochs from 400 to 500 the
WorldView-3 reduced-resolution PSNR is 39.458, 39.465, 39.458, 39.481, 39.476,
39.479, 39.502, 39.464, 39.496, 39.463 and 39.481~dB, and every value of that band
still leads FusionMamba by at least 0.08~dB. Baseline outputs are those distributed
with PanCollection and are scored with the same toolbox, except AWLP, which we ran
here with that toolbox, and ADWM, whose rows are quoted from its paper.

\begin{table*}[!t]
\centering
\caption{Reduced-resolution comparison on PanCollection (20 test images). PSNR in dB, SAM in degrees, ERGAS dimensionless; higher is better for PSNR and $Q$, lower for SAM and ERGAS. Best in bold, second best underlined.}
\label{tab:rr}
\footnotesize
\setlength{\tabcolsep}{4pt}
\renewcommand{\arraystretch}{0.94}
\begin{tabular}{lcccccccccccc}
\toprule
\multirow{2}{*}{Method} & \multicolumn{4}{c}{WorldView-3} & \multicolumn{4}{c}{GF2} & \multicolumn{4}{c}{QuickBird}\\
\cmidrule(lr){2-5}\cmidrule(lr){6-9}\cmidrule(lr){10-13}
 & PSNR & SAM & ERGAS & $Q_8$ & PSNR & SAM & ERGAS & $Q_4$ & PSNR & SAM & ERGAS & $Q_4$\\
\midrule
TV & 32.381 & 5.692 & 4.855 & 0.795 & 41.262 & 1.911 & 1.737 & 0.907 & 32.136 & 7.510 & 7.690 & 0.821\\
AWLP & 32.857 & 5.230 & 4.660 & 0.835 & 41.031 & 1.938 & 1.637 & 0.873 & 32.485 & 8.189 & 7.577 & 0.829\\
MTF-GLP-FS & 32.963 & 5.316 & 4.700 & 0.833 & 41.565 & 1.655 & 1.589 & 0.897 & 32.709 & 7.792 & 7.373 & 0.835\\
BDSD-PC & 32.970 & 5.428 & 4.697 & 0.829 & 41.205 & 1.681 & 1.667 & 0.892 & 32.550 & 8.085 & 7.513 & 0.831\\
PNN & 37.313 & 3.677 & 2.681 & 0.893 & 45.096 & 1.048 & 1.057 & 0.960 & 36.942 & 5.181 & 4.468 & 0.918\\
PanNet & 37.346 & 3.613 & 2.664 & 0.891 & 46.268 & 0.997 & 0.919 & 0.967 & 34.678 & 5.767 & 5.859 & 0.885\\
DiCNN & 37.390 & 3.592 & 2.672 & 0.900 & 44.931 & 1.053 & 1.081 & 0.959 & 35.781 & 5.367 & 5.133 & 0.904\\
FusionNet & 38.047 & 3.324 & 2.465 & 0.904 & 45.663 & 0.974 & 0.988 & 0.964 & 37.540 & 4.904 & 4.156 & 0.925\\
LAGConv & 38.568 & 3.104 & 2.300 & 0.910 & 48.760 & 0.786 & 0.687 & 0.980 & 38.181 & 4.547 & 3.826 & 0.934\\
DCFNet & 38.953 & 3.038 & 2.165 & 0.913 & 47.565 & 0.872 & 0.784 & 0.974 & 38.228 & 4.512 & 3.809 & 0.934\\
CANNet & 39.003 & 2.941 & 2.174 & 0.920 & 49.520 & 0.707 & 0.630 & 0.983 & \underline{38.488} & 4.496 & \underline{3.698} & \underline{0.937}\\
Pan-Mamba & 39.012 & 2.913 & 2.184 & 0.920 & 48.931 & 0.743 & 0.684 & 0.982 & 37.356 & 4.625 & 4.277 & 0.929\\
U2Net & 39.117 & 2.888 & 2.150 & 0.920 & 49.404 & 0.714 & 0.632 & 0.982 & 38.065 & 4.642 & 3.987 & 0.931\\
ADWM & 39.170 & 2.913 & 2.145 & 0.921 & \underline{49.909} & \underline{0.672} & \underline{0.597} & \best{0.985} & 38.466 & \underline{4.450} & 3.705 & \underline{0.937}\\
FusionMamba & \underline{39.374} & \underline{2.844} & \underline{2.092} & \underline{0.922} & 49.678 & 0.705 & 0.615 & \underline{0.984} & 37.986 & 4.610 & 4.054 & 0.930\\
\textbf{S2WMamba} & \best{39.502} & \best{2.808} & \best{2.054} & \best{0.923} & \best{50.083} & \best{0.668} & \best{0.585} & \best{0.985} & \best{38.802} & \best{4.381} & \best{3.565} & \best{0.939}\\
\bottomrule
\end{tabular}
\end{table*}

\begin{figure*}[!t]
\centering
\includegraphics[height=0.59\textheight,keepaspectratio]{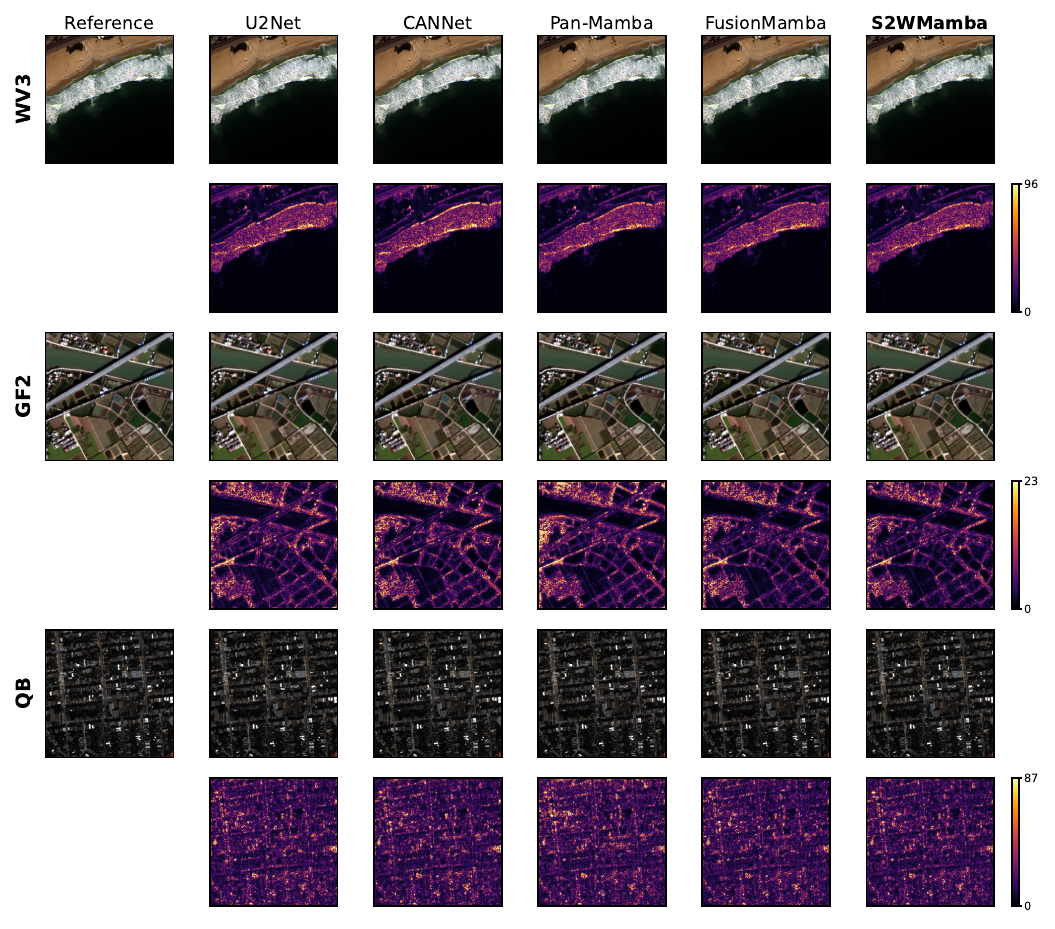}
\caption{Reduced-resolution results on the three sensors: the reference, then four
strong baselines and S2WMamba, each with its absolute-error map on a per-sensor scale
printed at the right (channel-mean error, 11-bit DN). Samples are the per-sensor
median by S2WMamba's PSNR.}
\label{fig:vis}
\end{figure*}

\begin{table*}[!t]
\centering
\caption{Full-resolution no-reference indices on the three sensors. Lower is better
for $D_\lambda$ and $D_s$, higher for HQNR. Best in bold, second best underlined.
HQNR is computed from the unrounded terms. All rows are rescored here under one
protocol from the outputs released with each method, except ADWM, which is quoted
from its paper, and DCFNet's WorldView-3 row, quoted from its own 50-image split and
set in italics because it is not comparable. S2WMamba reports the checkpoint with
the best test-split PSNR of its run.}
\label{tab:fr}
\footnotesize
\setlength{\tabcolsep}{4pt}
\begin{tabular}{lccccccccc}
\toprule
 & \multicolumn{3}{c}{WorldView-3} & \multicolumn{3}{c}{GF2} & \multicolumn{3}{c}{QuickBird}\\
\cmidrule(lr){2-4}\cmidrule(lr){5-7}\cmidrule(lr){8-10}
Method & $D_\lambda$ & $D_s$ & HQNR & $D_\lambda$ & $D_s$ & HQNR & $D_\lambda$ & $D_s$ & HQNR\\
\midrule
LAGConv & 0.037 & 0.042 & 0.923 & 0.028 & 0.079 & 0.895 & 0.087 & 0.067 & 0.852\\
AWLP & \best{0.016} & 0.050 & 0.935 & 0.037 & 0.133 & 0.835 & \underline{0.042} & 0.138 & 0.827\\
DCFNet & \emph{0.022} & \emph{0.022} & \emph{0.956} & 0.024 & 0.066 & 0.912 & 0.047 & 0.124 & 0.835\\
PanNet & \underline{0.017} & 0.047 & 0.937 & \underline{0.018} & 0.080 & 0.904 & 0.043 & 0.114 & 0.849\\
PNN & 0.021 & 0.043 & 0.937 & 0.032 & 0.094 & 0.877 & 0.058 & 0.062 & 0.884\\
FusionNet & 0.024 & 0.036 & 0.941 & 0.035 & 0.101 & 0.867 & 0.057 & 0.052 & 0.894\\
ADWM & 0.019 & \best{0.026} & \best{0.955} & 0.021 & 0.049 & 0.931 & 0.054 & \underline{0.029} & \best{0.919}\\
CANNet & 0.020 & 0.030 & 0.951 & 0.019 & 0.063 & 0.919 & \best{0.037} & 0.050 & 0.915\\
U2Net & 0.018 & 0.031 & 0.951 & 0.024 & 0.051 & 0.927 & 0.059 & \best{0.026} & \underline{0.916}\\
Pan-Mamba & 0.018 & 0.031 & 0.952 & 0.023 & 0.057 & 0.921 & 0.049 & 0.044 & 0.910\\
FusionMamba & 0.019 & \underline{0.027} & \best{0.955} & \best{0.017} & \best{0.030} & \best{0.954} & 0.057 & 0.040 & 0.906\\
\textbf{S2WMamba} & 0.021 & \best{0.026} & \underline{0.954} & 0.019 & \underline{0.039} & \underline{0.943} & 0.056 & 0.031 & 0.914\\
\bottomrule
\end{tabular}
\end{table*}

\begin{figure*}[!t]
\centering
\includegraphics[height=0.552\textheight,keepaspectratio]{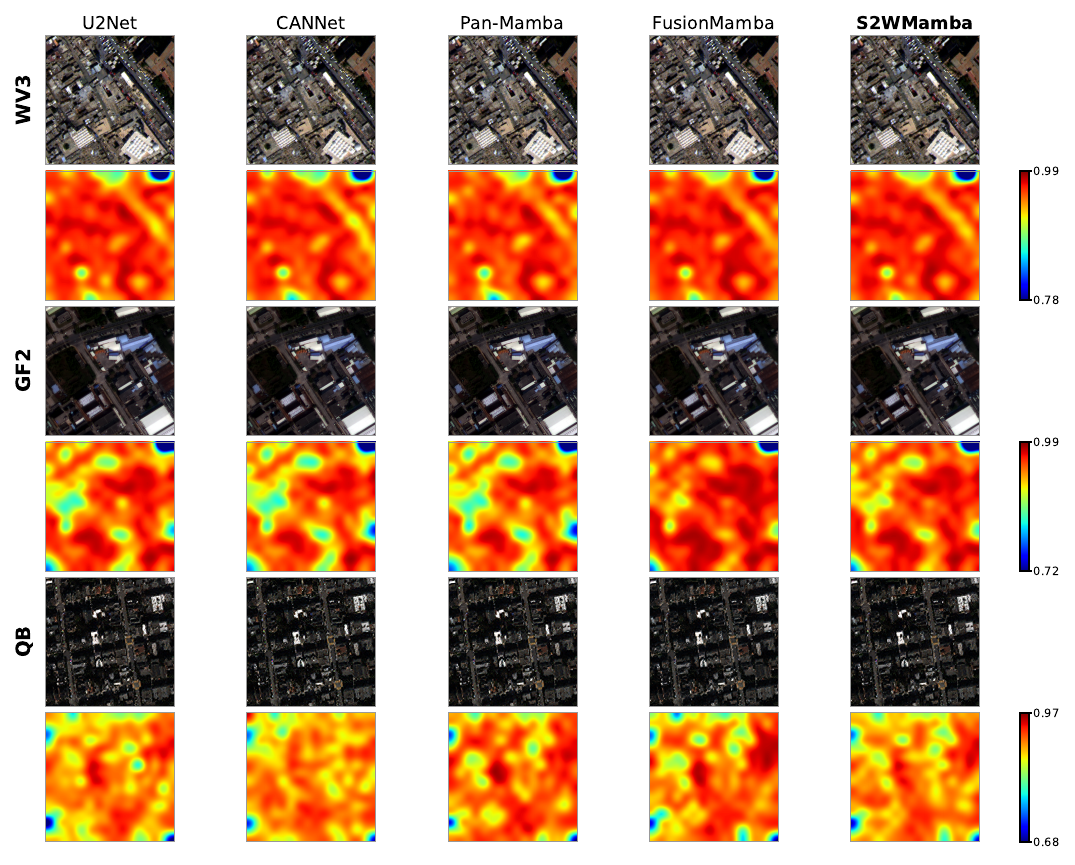}
\caption{Block-wise HQNR on the three sensors, one column per method and two rows
per sensor: the fused product above its block-wise HQNR map, computed with the same
definition as Table~\ref{tab:fr}: $D_\lambda$
from the block-wise $Q_{2n}$ against the MTF-filtered fusion product, $D_s$ from
the per-block difference of the two UQI terms. For each sensor the map row shares
one colour scale, whose endpoints are printed at its right; red is high. The
sample shown for each sensor is the median by
S2WMamba's HQNR over the 20 test images. The maps are a local diagnostic: a
per-block value can only carry the local $|Q_{\text{high}}-Q_{\text{low}}|$, while
the reported $D_s$ averages the two terms before differencing, so the block grid
mean is not the scalar HQNR of Table~\ref{tab:fr}.}
\label{fig:hqnr}
\end{figure*}

\subsection{Reduced-Resolution Results}
Table~\ref{tab:rr} compares S2WMamba with fifteen methods. S2WMamba is best on all four indices for
each of the three sensors. ADWM's GF2 row needed one correction: in its table the
eight baselines we share sit 6.025~dB below ours in PSNR while their SAM, ERGAS and
$Q_4$ agree to the last digit, so its GF2 PSNR is quoted here shifted by 6.025~dB.

Fig.~\ref{fig:vis} shows one test image per sensor fused by four baselines and by
S2WMamba, each followed by its absolute error against the reference. The error is
concentrated on building edges and along the shoreline, i.e. where the methods
disagree about how much PAN detail to inject, and for every sensor the S2WMamba
panels carry less of it than the four baselines shown.

The margin over the strongest baseline is 0.128~dB on WorldView-3 (FusionMamba),
0.174~dB on GF2 (ADWM) and 0.314~dB on QuickBird (CANNet).

DCFNet and LAGConv, which change the fusion operator rather than the decomposition,
sit 0.55--0.93~dB below it.

\subsection{Full-Resolution Results}
Table~\ref{tab:fr} reports the no-reference indices computed directly from each
method's output, except ADWM's, which are quoted from its paper, and DCFNet's
WorldView-3 row, quoted from its own 50-image split and not ranked here; there it
reports an HQNR of 0.956, above every row rescored in this table.
S2WMamba ranks third on WorldView-3 (0.954, behind FusionMamba and
ADWM at 0.955) and second on GF2 (0.943, behind FusionMamba's 0.954), and fourth on
QuickBird (0.914, behind ADWM's 0.919, U2Net's 0.916 and CANNet's 0.915); on every
sensor the best and ours
are no more than 0.011 apart.

On GF2 the spectral term of S2WMamba ($D_\lambda=0.019$) is close to the best in
the field, FusionMamba at 0.017, and the two differ in the spatial term: 0.039
against 0.030. $D_\lambda$ low-pass filters the product with the sensor MTF and compares
it with the interpolated MS image, so it is blind to the high-frequency content; $D_s$ compares
the agreement a fused band reaches with a resolution-matched PAN against that of the
interpolated MS image, and is minimised when the two coincide, so under- and
over-injection are penalised alike. Neither term refers to a reference, and both
are averages over $32\times32$ blocks, so a few blocks decide a rank and
differences of a few thousandths separate methods that are visually
indistinguishable. What the ordering in Table~\ref{tab:fr} measures is therefore
consistency with the statistics of the inputs, not fidelity to the reference --- on
GF2, where the reduced-resolution protocol does have a reference, the same model
leads its nearest competitor by 0.174~dB. Our own HQNR moves by at most 0.002 over cut sizes of 256 and 512 pixels and
paddings from 0 to 256, on GF2 and QuickBird.

Fig.~\ref{fig:hqnr} shows the local HQNR of five representative methods on the
three sensors, one value per $32\times32$ block. The low blocks of all methods
cluster on the same kind of structures that the error maps of
Fig.~\ref{fig:vis} highlight --- building edges and shorelines --- which is where
the index is lost. The block grid mean is a different statistic from the scalar
index of Table~\ref{tab:fr}, so the figure is read as a map of where quality is
lost, not as a ranking.

\subsection{Ablation Study}
\label{sec:ablation}
\begin{table}[!ht]
\centering
\caption{Ablation on WorldView-3 reduced resolution. All rows are trained for 500
epochs with the same recipe; the variant differs only in the component under test.
PSNR in dB, SAM in degrees, ERGAS dimensionless; higher is better for PSNR and
$Q_8$, lower for SAM and ERGAS. Best in bold.}
\label{tab:abl}
\begin{tabular}{lcccc}
\toprule
Variant & PSNR & SAM & ERGAS & $Q_8$\\
\midrule
S2WMamba (proposed)                    & \best{39.502} & \best{2.808} & \best{2.054} & \best{0.9226}\\
Fixed Haar basis                       & 39.378 & 2.857 & 2.088 & 0.9209\\
No spectral branch                     & 39.362 & 2.856 & 2.091 & 0.9212\\
Fully learned basis                    & 39.354 & 2.852 & 2.094 & 0.9218\\
Learned detail, unit norm              & 39.348 & 2.853 & 2.096 & 0.9207\\
Mamba $\rightarrow$ convolution        & 39.292 & 2.854 & 2.107 & 0.9204\\
Parallel branches                      & 39.266 & 2.887 & 2.118 & 0.9208\\
Reversed cascade                       & 39.265 & 2.900 & 2.118 & 0.9204\\
Aggregator $\rightarrow$ sum           & 39.252 & 2.894 & 2.129 & 0.9215\\
No low/high mixing (spectral)          & 39.228 & 2.886 & 2.120 & 0.9193\\
No spatial branch                      & 38.724 & 2.970 & 2.258 & 0.9169\\
\bottomrule
\end{tabular}
\end{table}

Table~\ref{tab:abl} reports the results; every row uses the identical recipe and the
same MATLAB implementation.

\emph{Wavelet basis.} Against the fixed Haar basis, freeing the whole basis lowers
PSNR by 0.024~dB and raises ERGAS by 0.006, while SAM and $Q_8$ come out
marginally better and the parameter count rises by 66k; constraining the detail filters to
unit norm lowers PSNR by 0.030~dB. The parameterisation of
Section~\ref{sec:lofix}, in which the approximation band is fixed and the detail
bands are learned, is 0.124~dB \emph{above} the fixed basis.

\begin{figure}[!tb]
\centering
\includegraphics[width=\columnwidth]{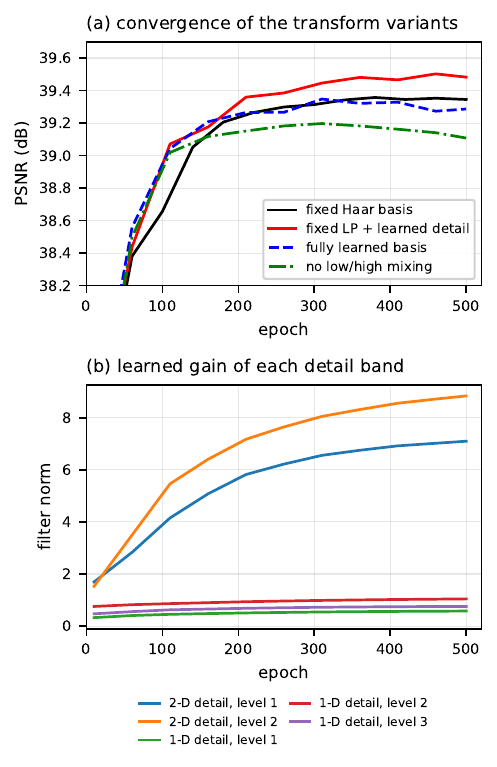}
\caption{(a) PSNR on the WorldView-3 test set against training epoch for four
parameterisations of the transform; every point is measured, not interpolated.
(b) Norm of each learnable detail filter over training. The approximation bands are
not plotted; they are fixed in
the fixed-Haar and proposed cases, learned in the fully learned case, and absent
from the no-mixing case.}
\label{fig:wavelet}
\end{figure}

Fig.~\ref{fig:wavelet} shows what separates the four cases. In
Fig.~\ref{fig:wavelet}(a) the two variants that either free the whole basis or
remove the mixing reach a plateau between 39.11 and 39.36~dB, both of them
\emph{declining} after epoch 400, whereas the proposed parameterisation is still
improving at epoch 460 and peaks 0.15~dB above the Haar curve at the same epoch. The
constrained transform therefore does not converge faster; it converges to a
better point, later, and the last 40 epochs cost it about 0.02~dB.

Fig.~\ref{fig:wavelet}(b) answers the obvious follow-up: what does the network do
with the freedom it is given. Each curve is the norm of one learnable detail filter
over training. The two 2-D filters are amplified several-fold, from 1.7 and 1.5 to
7.1 and 8.8, whereas the three 1-D filters move by less than a factor of two, from
0.32, 0.75 and 0.47 to 0.57, 1.04 and 0.75; the five filters therefore end at five
different scales. Which of the two axes uses that freedom more heavily is a property
of the trained solution, not of the design, since both are parameterised the same way.
This is why fixing every detail filter to unit norm costs 0.154~dB relative to
the proposed parameterisation: the constraint removes the quantity the model is
actually using. The relation is non-monotonic in adaptivity, which is why we
attribute the gain to per-level gain freedom rather than to either fixing the
basis or fixing the filter norm.

\emph{Replacing the mixing step.} With the Haar model as the reference, replacing the
two-point low/high mixing by a plain split of the channel pairs costs 0.150~dB on the
spectral axis. The variant is in fact worse than removing the spectral branch
altogether (39.228 against 39.362~dB), so the channel pairing on its own is not a
substitute for the mixing: what the two halves carry, not only how they are formed,
is what the branch uses.

\emph{Cascade direction.} Running the branches in parallel costs 0.236~dB.
Reversing the cascade --- spectral branch first, spatial branch conditioned on it
--- costs 0.237~dB, i.e. it is indistinguishable from no coupling at all in this
single run (0.001~dB apart). This is the sharpest result in the table, and it rules out the
two explanations that would otherwise be available. It is not that any second
branch adds capacity, since the reversed variant has exactly the parameters of the
main model and still performs as the uncoupled one. Nor is it that the spectral
branch simply needs more input, since it receives strictly more information in the
main model (\emph{its own} PAN features plus the spatial branch's estimate) than in
the reversed one. What the reversed variant lacks is a route by which the spatial
estimate reaches the spectral branch \emph{before} the channel sub-bands are formed,
and that is the component whose removal coincides with the whole loss.

\emph{Branches and operator.} Removing the spatial branch costs 0.778~dB, by far
the largest drop in the table; removing the spectral branch costs 0.140~dB.
Replacing the Mamba blocks by convolutional blocks of matched parameter count costs
0.210~dB, so the fusion stage contributes global modelling rather than capacity;
replacing the aggregator by a plain sum costs 0.250~dB, at 145k fewer parameters
(614,386 against 759,463). The asymmetry between the
branches is itself informative: the spatial branch carries the MS content the output
is built on, whereas the spectral branch only modulates how the PAN detail is
distributed.

\subsection{Efficiency}
\begin{table}[!ht]
\centering
\caption{Cost of one forward pass on an RTX 5090.}
\label{tab:cost}
\footnotesize
\setlength{\tabcolsep}{4pt}
\begin{tabular}{lccccc}
\toprule
Input & Params & Res. & GMACs & Latency\\
\midrule
\multirow{2}{*}{WV3} & \multirow{2}{*}{759,463} & $256^2$ & 33.1  & 41.0\,ms\\
                     &                          & $512^2$ & 132.3 & 163.9\,ms\\
\multirow{2}{*}{GF2/QB} & \multirow{2}{*}{579,875} & $256^2$ & 22.0  & 35.2\,ms\\
                        &                          & $512^2$ & 87.9  & 126.8\,ms\\
\bottomrule
\end{tabular}
\end{table}

Table~\ref{tab:cost} lists the parameter count and the cost of a single forward
pass measured on an RTX~5090.

\section{Conclusion}
We have presented S2WMamba, whose decomposition is matched to the modality: a 2-D
wavelet on an embedding of the PAN image and a 1-D transform along the spectral axis
of the fused cue, with the approximation band fixed and only the detail bands
learned. The branches are cascaded in one tested direction; reversing it removes the
entire benefit, which locates the contribution in the direction of the coupling
rather than in its presence. The ablation also fixes where the transform contributes:
freeing the whole basis costs 0.024~dB and fixing the detail norms costs 0.030~dB,
so the useful freedom is the per-level gain of the detail filters, not the basis. The method is best on all four reduced-resolution
indices on all three sensors; by HQNR it is third on WorldView-3, second on GF2 and
fourth on QuickBird, within 0.011 of the best. The reported model uses one
checkpoint per sensor, chosen on the reduced-resolution test split, and the same
weights serve both protocols; the choice moves the headline by at most 0.044~dB.
Two limitations remain: by HQNR we
trail the strongest baseline by 0.001--0.011 on every sensor, and the 1-D transform
assumes a known band order (permuting it costs 6--10~dB). The two transforms are
the only modality-specific parts of the network: the fusion stages are one
cross-modal scan, shared by both branches and by the aggregator, so the design
carries over to a new sensor by changing what is decomposed rather than what fuses
it. The natural next step is on the 1-D side, where a sensor with a non-dyadic band
count, or with bands that are not adjacent in wavelength, would need the pairing
itself to be learned rather than assumed. Parallel branches, the reversed cascade
and a plain-sum aggregator all land 0.24--0.25~dB below the model.

\renewcommand{\IEEEbibitemsep}{-1.5pt}
\bibliographystyle{IEEEtran}
\bibliography{refs}

@article{vivone2015critical,
  title = {A Critical Comparison Among Pansharpening Algorithms},
  author = {Vivone, Gemine and Alparone, Luciano and Chanussot, Jocelyn and Dalla Mura, Mauro and Garzelli, Andrea and Licciardi, Giorgio A. and Restaino, Rocco and Wald, Lucien},
  journal = {IEEE Transactions on Geoscience and Remote Sensing},
  volume = {53},
  number = {5},
  pages = {2565--2586},
  year = {2015},
  publisher = {Institute of Electrical and Electronics Engineers (IEEE)},
  doi = {10.1109/tgrs.2014.2361734}
}

@article{vivone2019component,
  title = {Robust Band-Dependent Spatial-Detail Approaches for Panchromatic Sharpening},
  author = {Vivone, Gemine},
  journal = {IEEE Transactions on Geoscience and Remote Sensing},
  volume = {57},
  number = {9},
  pages = {6421--6433},
  year = {2019},
  publisher = {Institute of Electrical and Electronics Engineers (IEEE)},
  doi = {10.1109/tgrs.2019.2906073}
}

@article{vivone2018multiresolution,
  title = {Full Scale Regression-Based Injection Coefficients for Panchromatic Sharpening},
  author = {Vivone, Gemine and Restaino, Rocco and Chanussot, Jocelyn},
  journal = {IEEE Transactions on Image Processing},
  volume = {27},
  number = {7},
  pages = {3418--3431},
  year = {2018},
  publisher = {Institute of Electrical and Electronics Engineers (IEEE)},
  doi = {10.1109/tip.2018.2819501}
}

@article{palsson2013variational,
  title = {A New Pansharpening Algorithm Based on Total Variation},
  author = {Palsson, Frosti and Sveinsson, Johannes R. and Ulfarsson, Magnus O.},
  journal = {IEEE Geoscience and Remote Sensing Letters},
  volume = {11},
  number = {1},
  pages = {318--322},
  year = {2014},
  publisher = {Institute of Electrical and Electronics Engineers (IEEE)},
  doi = {10.1109/lgrs.2013.2257669}
}

@article{deng2022machine,
  title = {Machine Learning in Pansharpening: A benchmark, from shallow to deep networks},
  author = {Deng, Liang-jian and Vivone, Gemine and Paoletti, Mercedes E. and Scarpa, Giuseppe and He, Jiang and Zhang, Yongjun and Chanussot, Jocelyn and Plaza, Antonio},
  journal = {IEEE Geoscience and Remote Sensing Magazine},
  volume = {10},
  number = {3},
  pages = {279--315},
  year = {2022},
  publisher = {Institute of Electrical and Electronics Engineers (IEEE)},
  doi = {10.1109/mgrs.2022.3187652}
}

@article{mallat1989theory,
  title = {A theory for multiresolution signal decomposition: the wavelet representation},
  author = {Mallat, S.G.},
  journal = {IEEE Transactions on Pattern Analysis and Machine Intelligence},
  volume = {11},
  number = {7},
  pages = {674--693},
  year = {1989},
  publisher = {Institute of Electrical and Electronics Engineers (IEEE)},
  doi = {10.1109/34.192463}
}

@article{masi2016pansharpening,
  title = {Pansharpening by Convolutional Neural Networks},
  author = {Masi, Giuseppe and Cozzolino, Davide and Verdoliva, Luisa and Scarpa, Giuseppe},
  journal = {Remote Sensing},
  volume = {8},
  number = {7},
  pages = {594},
  year = {2016},
  publisher = {MDPI AG},
  doi = {10.3390/rs8070594}
}

@inproceedings{yang2017pannet,
  title = {PanNet: A Deep Network Architecture for Pan-Sharpening},
  author = {Yang, Junfeng and Fu, Xueyang and Hu, Yuwen and Huang, Yue and Ding, Xinghao and Paisley, John},
  booktitle = {2017 IEEE International Conference on Computer Vision (ICCV)},
  pages = {1753--1761},
  year = {2017},
  publisher = {IEEE},
  doi = {10.1109/iccv.2017.193}
}

@article{he2019dicnn,
  title = {Pansharpening via Detail Injection Based Convolutional Neural Networks},
  author = {He, Lin and Rao, Yizhou and Li, Jun and Chanussot, Jocelyn and Plaza, Antonio and Zhu, Jiawei and Li, Bo},
  journal = {IEEE Journal of Selected Topics in Applied Earth Observations and Remote Sensing},
  volume = {12},
  number = {4},
  pages = {1188--1204},
  year = {2019},
  publisher = {Institute of Electrical and Electronics Engineers (IEEE)},
  doi = {10.1109/jstars.2019.2898574}
}

@article{deng2021fusionnet,
  title = {Detail Injection-Based Deep Convolutional Neural Networks for Pansharpening},
  author = {Deng, Liang-Jian and Vivone, Gemine and Jin, Cheng and Chanussot, Jocelyn},
  journal = {IEEE Transactions on Geoscience and Remote Sensing},
  volume = {59},
  number = {8},
  pages = {6995--7010},
  year = {2021},
  publisher = {Institute of Electrical and Electronics Engineers (IEEE)},
  doi = {10.1109/tgrs.2020.3031366}
}

@inproceedings{duan2024cannet,
  title = {Content-Adaptive Non-Local Convolution for Remote Sensing Pansharpening},
  author = {Duan, Yule and Wu, Xiao and Deng, Haoyu and Deng, Liang-Jian},
  booktitle = {2024 IEEE/CVF Conference on Computer Vision and Pattern Recognition (CVPR)},
  pages = {27738--27747},
  year = {2024},
  publisher = {IEEE},
  doi = {10.1109/cvpr52733.2024.02620}
}

@inproceedings{zhou2022panformer,
  title = {PanFormer: A Transformer Based Model for Pan-Sharpening},
  author = {Zhou, Huanyu and Liu, Qingjie and Wang, Yunhong},
  booktitle = {2022 IEEE International Conference on Multimedia and Expo (ICME)},
  pages = {1--6},
  year = {2022},
  publisher = {IEEE},
  doi = {10.1109/icme52920.2022.9859770}
}

@inproceedings{peng2023u2net,
  title = {U2Net: A General Framework with Spatial-Spectral-Integrated Double U-Net for Image Fusion},
  author = {Peng, Siran and Guo, Chenhao and Wu, Xiao and Deng, Liang-Jian},
  booktitle = {Proceedings of the 31st ACM International Conference on Multimedia},
  pages = {3219--3227},
  year = {2023},
  publisher = {ACM},
  doi = {10.1145/3581783.3612084}
}

@article{peng2024fusionmamba,
  title = {FusionMamba: Efficient Remote Sensing Image Fusion With State Space Model},
  author = {Peng, Siran and Zhu, Xiangyu and Deng, Haoyu and Deng, Liang-Jian and Lei, Zhen},
  journal = {IEEE Transactions on Geoscience and Remote Sensing},
  volume = {62},
  pages = {1--16},
  year = {2024},
  publisher = {Institute of Electrical and Electronics Engineers (IEEE)},
  doi = {10.1109/tgrs.2024.3496073}
}

@inproceedings{li2025quaternion,
  title = {Enhanced Pansharpening Via Quaternion Spatial-Spectral Interactions},
  author = {Li, Dong and Luo, Chunhui and Bao, Yuanfei and Yang, Gang and Xiao, Jie and Fu, Xueyang and Zha, Zheng-Jun},
  booktitle = {2025 IEEE/CVF International Conference on Computer Vision (ICCV)},
  pages = {10908--10918},
  year = {2025},
  publisher = {IEEE},
  doi = {10.1109/iccv51701.2025.01015}
}

@article{yang2024sffnet,
  title = {SFFNet: A Wavelet-Based Spatial and Frequency Domain Fusion Network for Remote Sensing Segmentation},
  author = {Yang, Yunsong and Yuan, Genji and Li, Jinjiang},
  journal = {IEEE Transactions on Geoscience and Remote Sensing},
  volume = {62},
  pages = {1--17},
  year = {2024},
  publisher = {Institute of Electrical and Electronics Engineers (IEEE)},
  doi = {10.1109/tgrs.2024.3427370}
}

@article{xing2024crossdiff,
  title = {CrossDiff: Exploring Self-Supervised Representation of Pansharpening via Cross-Predictive Diffusion Model},
  author = {Xing, Yinghui and Qu, Litao and Zhang, Shizhou and Zhang, Kai and Zhang, Yanning and Bruzzone, Lorenzo},
  journal = {IEEE Transactions on Image Processing},
  volume = {33},
  pages = {5496--5509},
  year = {2024},
  publisher = {Institute of Electrical and Electronics Engineers (IEEE)},
  doi = {10.1109/tip.2024.3461476}
}

@article{arienzo2022full,
  title = {Full-Resolution Quality Assessment of Pansharpening: Theoretical and hands-on approaches},
  author = {Arienzo, Alberto and Vivone, Gemine and Garzelli, Andrea and Alparone, Luciano and Chanussot, Jocelyn},
  journal = {IEEE Geoscience and Remote Sensing Magazine},
  volume = {10},
  number = {3},
  pages = {168--201},
  year = {2022},
  publisher = {Institute of Electrical and Electronics Engineers (IEEE)},
  doi = {10.1109/mgrs.2022.3170092}
}

@article{he2024panmamba,
  title = {Pan-Mamba: Effective pan-sharpening with state space model},
  author = {He, Xuanhua and Cao, Ke and Zhang, Jie and Yan, Keyu and Wang, Yingying and Li, Rui and Xie, Chengjun and Hong, Danfeng and Zhou, Man},
  journal = {Information Fusion},
  volume = {115},
  pages = {102779},
  year = {2025},
  publisher = {Elsevier BV},
  doi = {10.1016/j.inffus.2024.102779}
}

@inproceedings{liu2024vmamba,
  title = {VMamba: Visual State Space Model},
  author = {Liu, Yue and Tian, Yunjie and Zhao, Yuzhong and Yu, Hongtian and Xie, Lingxi and Wang, Yaowei and Ye, Qixiang and Jiao, Jianbin and Liu, Yunfan},
  booktitle = {Advances in Neural Information Processing Systems 37},
  pages = {103031--103063},
  year = {2024},
  publisher = {Neural Information Processing Systems Foundation, Inc. (NeurIPS)},
  doi = {10.52202/079017-3273}
}

@inproceedings{zhong2024ssdiff,
  title = {SSDiff: Spatial-spectral Integrated Diffusion Model for Remote Sensing Pansharpening},
  author = {Zhong, Yu and Wu, Xiao and Cao, Zihan and Dou, Hong-Xia and Deng, Liang-Jian},
  booktitle = {Advances in Neural Information Processing Systems 37},
  pages = {77962--77986},
  year = {2024},
  publisher = {Neural Information Processing Systems Foundation, Inc. (NeurIPS)},
  doi = {10.52202/079017-2478}
}

@article{zhao2017loss,
  title = {Loss Functions for Image Restoration With Neural Networks},
  author = {Zhao, Hang and Gallo, Orazio and Frosio, Iuri and Kautz, Jan},
  journal = {IEEE Transactions on Computational Imaging},
  volume = {3},
  number = {1},
  pages = {47--57},
  year = {2017},
  publisher = {Institute of Electrical and Electronics Engineers (IEEE)},
  doi = {10.1109/tci.2016.2644865}
}

@article{wald1997fusion,
  title = {Fusion of satellite images of different spatial resolutions: Assessing the quality of resulting images},
  author = {Wald, Lucien and Ranchin, Thierry and Mangolini, Marc},
  journal = {Photogrammetric Engineering and Remote Sensing},
  volume = {63},
  number = {6},
  pages = {691--699},
  year = {1997}
}

@book{wald2002data,
  title = {Data Fusion: Definitions and Architectures: Fusion of Images of Different Spatial Resolutions},
  author = {Wald, Lucien},
  publisher = {Les Presses de l'{\'E}cole des Mines},
  address = {Paris, France},
  year = {2002}
}

@inproceedings{boardman1993automated,
  title = {Automated spectral analysis: A geological example using AVIRIS data, north Grapevine Mountains, Nevada},
  author = {Boardman, Joseph W. and Kruse, Fred A.},
  booktitle = {Proceedings of the Tenth Thematic Conference on Geologic Remote Sensing},
  address = {San Antonio, Texas, USA},
  pages = {407--418},
  year = {1994}
}

@inproceedings{gu2023mamba,
  title = {Mamba: Linear-Time Sequence Modeling with Selective State Spaces},
  author = {Gu, Albert and Dao, Tri},
  booktitle = {First Conference on Language Modeling (COLM)},
  year = {2024},
  note = {arXiv:2312.00752}
}

@inproceedings{loshchilov2017decoupled,
  title = {Decoupled Weight Decay Regularization},
  author = {Loshchilov, Ilya and Hutter, Frank},
  booktitle = {International Conference on Learning Representations},
  year = {2019},
  note = {arXiv:1711.05101}
}

@inproceedings{cui2024adair,
  title = {AdaIR: Adaptive All-in-One Image Restoration via Frequency Mining and Modulation},
  author = {Cui, Yuning and Zamir, Syed Waqas and Khan, Salman and Knoll, Alois and Shah, Mubarak and Khan, Fahad Shahbaz},
  booktitle = {International Conference on Learning Representations},
  year = {2025},
  note = {arXiv:2403.14614}
}

@inproceedings{yuhas1992discrimination,
  title = {Discrimination among semi-arid landscape endmembers using the Spectral Angle Mapper (SAM) algorithm},
  author = {Yuhas, Roberta H. and Goetz, Alexander F. H. and Boardman, Joe W.},
  booktitle = {Summaries of the 3rd Annual JPL Airborne Geoscience Workshop},
  pages = {147--149},
  year = {1992},
  address = {Pasadena, CA, USA}
}

@article{baty2024mamba2d,
  title = {Mamba2D: A Natively Multi-Dimensional State-Space Model for Vision Tasks},
  author = {Baty, Enis and Hern{\'a}ndez D{\'i}az, Alejandro and Davidson, Rebecca and Bridges, Chris and Hadfield, Simon},
  journal = {arXiv preprint arXiv:2412.16146v3},
  year = {2024}
}

@inproceedings{huang2025adwm,
  title     = {A General Adaptive Dual-Level Weighting Mechanism for Remote Sensing Pansharpening},
  author    = {Huang, Jie and others},
  booktitle = {Proc. IEEE/CVF Conf. Comput. Vis. Pattern Recognit. (CVPR)},
  year      = {2025},
  note      = {arXiv:2503.13214}
}

@article{alparone2008qnr,
  title   = {Multispectral and Panchromatic Data Fusion Assessment Without Reference},
  author  = {Alparone, Luciano and Aiazzi, Bruno and Baronti, Stefano and Garzelli, Andrea and Nencini, Filippo and Selva, Massimo},
  journal = {Photogrammetric Engineering \& Remote Sensing},
  volume  = {74},
  number  = {2},
  pages   = {193--200},
  year    = {2008}
}

@inproceedings{wu2021dcfnet,
  title     = {Dynamic Cross Feature Fusion for Remote Sensing Pansharpening},
  author    = {Wu, Xiao and Huang, Ting-Zhu and Deng, Liang-Jian and Zhang, Tian-Jing},
  booktitle = {Proc. IEEE/CVF Int. Conf. Comput. Vis. (ICCV)},
  pages     = {14667--14676},
  year      = {2021},
  doi       = {10.1109/ICCV48922.2021.01442}
}

@inproceedings{jin2022lagconv,
  title     = {{LAGConv}: Local-Context Adaptive Convolution Kernels with Global Harmonic Bias for Pansharpening},
  author    = {Jin, Zi-Rong and Zhang, Tian-Jing and Jiang, Tai-Xiang and Vivone, Gemine and Deng, Liang-Jian},
  booktitle = {Proc. AAAI Conf. Artif. Intell.},
  volume    = {36},
  number    = {1},
  pages     = {1113--1121},
  year      = {2022},
  doi       = {10.1609/aaai.v36i1.19996}
}

@article{meng2022vit,
  title   = {Vision Transformer for Pansharpening},
  author  = {Meng, Xiangchao and Wang, Nan and Shao, Feng and Li, Shutao},
  journal = {IEEE Trans. Geosci. Remote Sens.},
  volume  = {60},
  pages   = {1--11},
  year    = {2022},
  doi     = {10.1109/TGRS.2022.3168465}
}

@article{quan2023dtempan,
  title   = {{DTEMPan}: Dual Texture-Edge Maintaining Transformer for Pansharpening},
  author  = {Quan, Wuzhou and Guo, Wen},
  journal = {IEEE Trans. Geosci. Remote Sens.},
  volume  = {61},
  pages   = {1--16},
  year    = {2023},
  doi     = {10.1109/TGRS.2023.3335363}
}

@article{zhao2023asym,
  title   = {Asymmetric Bidirectional Fusion Network for Remote Sensing Pansharpening},
  author  = {Zhao, Xin and Guo, Jiayi and Zhang, Yueting and Wu, Yirong},
  journal = {IEEE Trans. Geosci. Remote Sens.},
  volume  = {61},
  pages   = {1--16},
  year    = {2023},
  doi     = {10.1109/TGRS.2023.3296510}
}

@article{wang2023duald,
  title   = {Dual-Domain Dynamic Local--Global Network for Pansharpening},
  author  = {Wang, Zeping and Hu, Jianwen and Feng, Xi and Kang, Xudong and Mo, Yan},
  journal = {IEEE Trans. Geosci. Remote Sens.},
  volume  = {61},
  pages   = {1--14},
  year    = {2023},
  doi     = {10.1109/TGRS.2023.3316892}
}

@article{zhang2023bp,
  title   = {Spatial--Spectral Dual Back-Projection Network for Pansharpening},
  author  = {Zhang, Kai and Wang, Anfei and Zhang, Feng and Wan, Wenbo and Sun, Jiande and Bruzzone, Lorenzo},
  journal = {IEEE Trans. Geosci. Remote Sens.},
  volume  = {61},
  pages   = {1--16},
  year    = {2023},
  doi     = {10.1109/TGRS.2023.3266799}
}

@article{rui2024zero,
  title   = {Variational Zero-Shot Multispectral Pansharpening},
  author  = {Rui, Xiangyu and Cao, Xiangyong and Li, Yining and Meng, Deyu},
  journal = {IEEE Trans. Geosci. Remote Sens.},
  volume  = {62},
  pages   = {1--16},
  year    = {2024},
  doi     = {10.1109/TGRS.2024.3492059}
}

@article{yang2023crossres,
  title   = {Cross-Resolution Semi-Supervised Adversarial Learning for Pansharpening},
  author  = {Yang, Guishuo and Zhang, Kai and Zhang, Feng and Wang, Jian and Sun, Jiande},
  journal = {IEEE Trans. Geosci. Remote Sens.},
  volume  = {61},
  pages   = {1--17},
  year    = {2023},
  doi     = {10.1109/TGRS.2023.3271658}
}

@article{chen2023aug,
  title   = {Spatial Data Augmentation: Improving the Generalization of Neural Networks for Pansharpening},
  author  = {Chen, Lihui and Vivone, Gemine and Nie, Zihao and Chanussot, Jocelyn and Yang, Xiaomin},
  journal = {IEEE Trans. Geosci. Remote Sens.},
  volume  = {61},
  pages   = {1--11},
  year    = {2023},
  doi     = {10.1109/TGRS.2023.3262262}
}

@article{zhang2022amf,
  title   = {Attention-Based Multistage Fusion Network for Remote Sensing Image Pansharpening},
  author  = {Zhang, Wanwan and Li, Jinjiang and Hua, Zhen},
  journal = {IEEE Trans. Geosci. Remote Sens.},
  volume  = {60},
  pages   = {1--16},
  year    = {2022},
  doi     = {10.1109/TGRS.2021.3113984}
}

@article{su2022trn,
  title   = {Transformer-Based Regression Network for Pansharpening Remote Sensing Images},
  author  = {Su, Xunyang and Li, Jinjiang and Hua, Zhen},
  journal = {IEEE Trans. Geosci. Remote Sens.},
  volume  = {60},
  pages   = {1--23},
  year    = {2022},
  doi     = {10.1109/TGRS.2022.3152425}
}

\end{document}